\PassOptionsToPackage{expansion=false,protrusion=false}{microtype}
\RequirePackage[bookmarksnumbered,unicode]{hyperref}
\documentclass[sigconf,nonacm]{cidr-2027}
\setcopyright{none}
\renewcommand\footnotetextcopyrightpermission[1]{}
\usepackage{booktabs}
\usepackage{placeins}

\begin{document}

\title{When Is Graph Structure Worth Its Cost?\\
The Case for Structure Pricing in Retrieval-Augmented Generation}

\author{Yuzhong Zhang}
\affiliation{%
  \institution{The Chinese University of Hong Kong, Shenzhen}
  \city{Shenzhen}
  \country{China}
}

\author{Haoyang Ma}
\affiliation{%
  \institution{The Hong Kong University of Science and Technology}
  \city{Hong Kong}
  \country{China}
}

\author{Chao Peng}
\affiliation{%
  \institution{University of Edinburgh}
  \city{Edinburgh}
  \country{United Kingdom}
}

\author{Lionel Briand}
\affiliation{%
  \institution{Lero, the Research Ireland Centre for Software, University of Limerick}
  \city{Limerick}
  \country{Ireland}
}
\affiliation{%
  \institution{University of Ottawa}
  \city{Ottawa}
  \country{Canada}
}
\email{lionel.briand@lero.ie}

\author{Boxi Yu}
\authornote{Corresponding author.}
\affiliation{%
  \institution{Lero, the Research Ireland Centre for Software, University of Limerick}
  \city{Limerick}
  \country{Ireland}
}
\email{boxi.yu@lero.ie}

\author{Jialun Cao}
\affiliation{%
  \institution{The Hong Kong University of Science and Technology}
  \city{Hong Kong}
  \country{China}
}

\begin{abstract}


Graph-based retrieval-augmented generation (RAG) can help answer questions that
require information from many documents. However, building a graph often
requires many language-model calls during ingestion. It is therefore important
to ask whether its quality gains justify the additional cost.

We present EffiRAG, a graph-based RAG system designed to reduce this cost. It
uses the graph to locate relevant passages and generates answers from the
original text. This design preserves source information while keeping graph
construction and query processing lightweight.

We evaluate EffiRAG on UltraDomain, which contains 120 open-ended questions
from four domains. Compared with LightRAG-hybrid, EffiRAG produces the preferred
answer on 93 questions. LightRAG is preferred on 7, and the remaining 20 are
splits. EffiRAG also reduces total system cost by 57\%, from \$0.952 to
\$0.408. The cost includes language-model calls during ingestion and querying.

The advantage remains as the corpus grows. At 10 and 20 documents per domain,
EffiRAG uses a lightweight, non-LLM filter to skip low-salience chunks. It
remains preferred over LightRAG-hybrid. It costs 4.2$\times$ and 4.5$\times$ less,
respectively.

The comparisons identify different quality--cost trade-offs.
Graph-based RAG systems should therefore be evaluated by both
answer quality and cost. The results favor graph structure that locates and
preserves source evidence.

\end{abstract}

\maketitle

\section{Introduction}

Retrieval-augmented generation (RAG)~\cite{rag} allows large language models to
answer questions over private or recently updated corpora without additional
fine-tuning. A standard RAG system splits documents into chunks, embeds them,
retrieves the chunks most relevant to a query, and generates an answer from the
retrieved text. This pipeline is simple and relatively inexpensive. However,
it can miss information distributed across several chunks, especially for
comparison, cross-document reasoning, or corpus synthesis.

Graph-based RAG systems address this limitation by representing entities and
relations explicitly. GraphRAG organizes graph elements into communities and
generates community summaries for corpus-level sensemaking~\cite{graphrag}.
LightRAG combines graph-based retrieval with vector retrieval~\cite{lightrag}.
These structures can connect information that chunk similarity alone may miss,
but extraction, relation discovery, summarization, and graph traversal add
language-model calls.

This paper therefore asks a practical question: \textbf{when does graph
structure improve answer quality enough to justify the cost of building and
using it?} We study this trade-off by measuring graph-construction and query
cost together with answer quality.

We introduce EffiRAG, a graph-based RAG system built around a simple principle:
the graph locates evidence, while source chunks supply the answer material.
During ingestion, EffiRAG extracts entities and relations from source chunks.
It links each extraction back to the chunk that supports it. During querying,
these extractions help locate relevant source chunks under a fixed context
budget. The generator answers from the original source text, with the
extractions serving as retrieval handles. We call this design
\emph{source-chunk grounding}.

EffiRAG controls ingestion cost through bounded graph extraction. It extracts
each source chunk once and omits community summarization, reducing ingestion
cost relative to LightRAG-hybrid.

We report \emph{system cost} as the input- and output-token charges for
provider-LLM calls during ingestion and querying. Evaluation judging is
excluded. Local embeddings are treated as unbilled.

In our main comparison, \emph{EffiRAG} applies this bounded extraction policy
to all source chunks and uses a fixed context budget. We evaluate it on
UltraDomain, which contains 120 open-ended queries across four domains.
EffiRAG is preferred over LightRAG-hybrid on 93 queries. LightRAG-hybrid is preferred on 7, and the remaining 20 produce no consistent preference. Additional checks show robustness to judging and answer-length effects. Across
ingestion and the 120 queries, EffiRAG costs \$0.408, compared with \$0.952 for
LightRAG-hybrid, a reduction of 57\%.

We also study larger corpora containing 10 and 20 documents per domain. EffiRAG
retains the same fixed-budget query policy. Before extraction, a lightweight,
non-LLM salience filter selects chunks likely to contain useful relations.
EffiRAG remains preferred over LightRAG-hybrid while reducing system cost by
factors of 4.2 and 4.5 in the 10- and 20-document-per-domain settings.

Under tight cost constraints, lightweight non-graph retrievers such as
BM25-vector~\cite{bm25}, NaiveRAG, and HyDE~\cite{hyde} may be more practical,
although they produce lower-quality answers on UltraDomain.

This paper makes three contributions:

\begin{enumerate}
\item We introduce EffiRAG, a source-grounded graph-RAG system. Extracted entities and relations guide retrieval. The original source chunks remain available to the answer generator. On UltraDomain, EffiRAG is preferred over LightRAG-hybrid at substantially lower system cost.

\item We provide a cost evaluation that jointly reports ingestion and
query-time provider-LLM usage.

\item We provide an empirical analysis of when graph construction is worthwhile. The analysis covers non-graph baselines, extraction-free graphs, larger corpora, and gold-answer multi-hop question answering.
\end{enumerate}

\section{Related Work}

\textbf{Graph construction.}
GraphRAG uses LLM-based entity and relation extraction, community detection,
and generated community summaries~\cite{graphrag}. LightRAG constructs an
entity--relation graph alongside vector indexes~\cite{lightrag}, while
LazyGraphRAG lowers indexing cost through non-LLM graph
construction~\cite{lazygraphrag}. EffiRAG retains bounded LLM extraction but
omits community summarization and links every extracted entity and relation to
its supporting source chunks.

\textbf{Graph use.}
GraphRAG retrieves community reports for corpus-level
questions~\cite{graphrag}. LightRAG combines graph and vector
retrieval~\cite{lightrag}, while PathRAG retrieves selected relational paths
to reduce irrelevant graph context~\cite{pathrag}. EffiRAG uses extracted
entities and relations to locate their supporting source chunks. The generator
receives the retrieved relations and original text, so the graph guides
retrieval while retaining source evidence.

\textbf{Cost--quality evaluation.}
GraphRAG-Bench studies whether the quality gains of graph-based RAG justify
its construction cost~\cite{graphragbench-when}, while LazyGraphRAG targets
lower indexing cost~\cite{lazygraphrag}. EffiRAG jointly measures
provider-LLM token cost during ingestion and querying and compares it with
judged answer quality. It also includes lower-cost non-graph retrievers to
identify when graph construction provides sufficient benefit.

\paragraph{Positioning.}
GraphRAG relies on community summaries for corpus-level synthesis.
LightRAG combines graph and vector retrieval.
LazyGraphRAG prioritizes low indexing cost via non-LLM construction.
EffiRAG takes a different route: it uses bounded LLM extraction to locate
original source chunks, and reports ingestion and querying cost jointly.

\section{Method}
\label{sec:method}

\begin{figure*}[t]
\centering
\includegraphics[width=\textwidth]{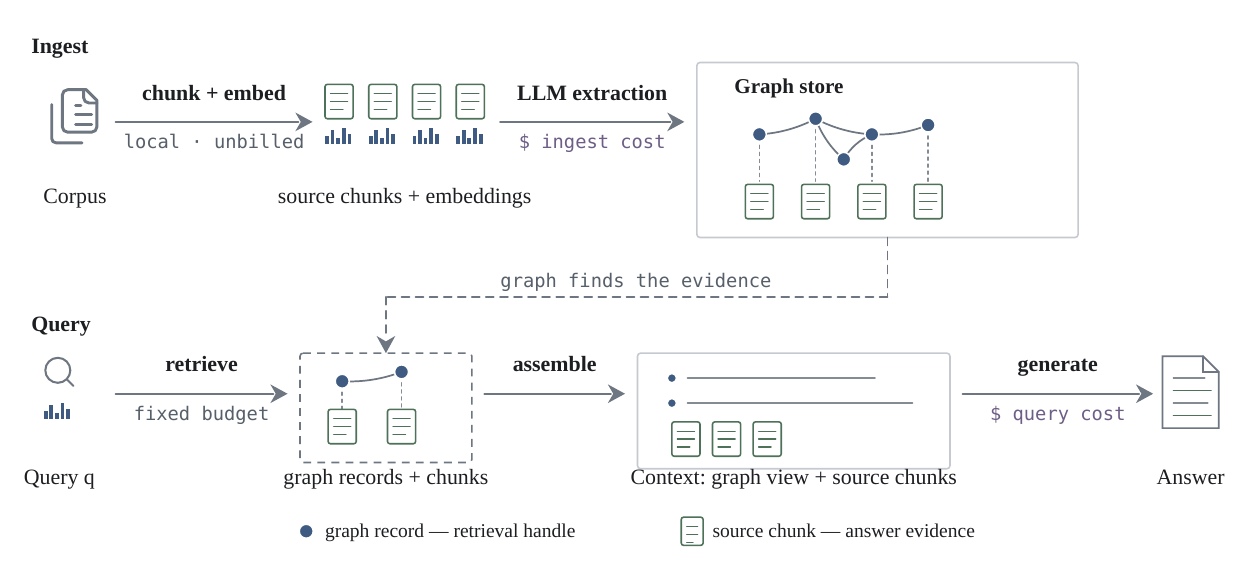}
\Description{EffiRAG pipeline. During ingestion, the corpus is divided into source chunks and embedded locally. An LLM extracts entities and relations and links them to their supporting chunks. During querying, EffiRAG retrieves entities, relations, and source chunks under a fixed context budget and generates an answer from the selected evidence.}
\caption{EffiRAG pipeline. Extracted entities and relations locate evidence; their supporting source chunks provide the original answer text.}
\label{fig:pipeline}
\end{figure*}

EffiRAG has two stages (Figure~\ref{fig:pipeline}).

Ingestion divides the corpus into source chunks. Each chunk is embedded locally.
An LLM extracts entities and relations from each chunk.

Querying retrieves relevant entities, relations, and source chunks. It selects
evidence under a fixed context budget. The generator then answers from the
selected evidence.

The central design principle is \emph{source-chunk grounding}: every
extracted entity or relation retains a link to the chunk that supports it.
The extracted entities and relations serve as retrieval handles, while the
source text remains the answer evidence. This matters because a compact relation may omit
information such as dates, negation, conditions, or uncertainty. EffiRAG gives
the generator both the retrieved relations and their supporting source chunks.

\paragraph{Ingestion.}
A \emph{source chunk} is a contiguous unit of original document text used for
extraction and retrieval. The graph store contains a source-chunk record for
each chunk and two types of \emph{graph record}: entity records and relation
records. A source-chunk record stores the original text and its embedding. An
entity record stores a normalized key, name, type, description, mention count,
and references to its supporting source chunks.
A relation record stores the source entity, target entity, relation keywords,
description, and a reference to its supporting source chunk.

For retrieval, each graph record is converted to \emph{record text}. Entity
record text has the form ``name (type): description.'' Relation record text
has the form ``source $\rightarrow$ target (keywords): description.'' The
graph store holds the records and their source links. Separate embedding
collections for graph-record text and source chunks support cosine-similarity
search.

\paragraph{Retrieval and generation.}
Given a query, EffiRAG embeds it and compares the query embedding with the
graph-record and source-chunk embeddings. Graph records are ranked by the
cosine similarity of their record-text embeddings. Source chunks are ranked
in the same way using their stored embeddings. EffiRAG follows the source links of the retrieved graph records. It combines
the resulting supporting chunks with those retrieved directly. Duplicates are
removed.

EffiRAG formats the selected graph records as a textual \emph{graph view}. The
final context contains this graph view together with the selected source
chunks. The graph view exposes entity and relation information, while the
source chunks provide the original answer evidence.

\paragraph{Query-time evidence selection.}
At query time, Coverage-Budgeted Evidence Selection (CBES) assembles graph
records and source chunks under a \emph{context budget} $B$. $B$ is the maximum
number of tokens placed in the final context.

Query aspects are named entities and noun phrases extracted from the query.
CBES uses a submodular coverage objective. An item receives less additional
value when its aspects are already covered by selected evidence. It receives
more value when it covers new aspects. It greedily
selects the item with the largest marginal coverage gain per token. It stops when no remaining item fits
the budget. This reduces redundant context.

\paragraph{Reported settings.}
EffiRAG combines source-chunk grounding, bounded graph extraction, and
fixed-budget evidence selection. Unless stated otherwise, \emph{EffiRAG}
refers to the default setting used in the five-document-per-domain comparison.

EffiRAG extracts every source chunk once and omits community summarization.
Before CBES, EffiRAG retrieves up to 16 entity records,
16 relation records, and 8 source chunks.
It then adds at most 4 supporting chunks.

These are candidate-pool caps, not the final context size.
The final context is separately bounded by $B=3000$.
The caps only need to be large enough to give CBES a diverse
candidate set. We chose 16/16/8/4 as a conservative setting
that is well above the typical number of items CBES selects.
We did not tune these values on UltraDomain.

\emph{EffiRAG w/ salience filtering} skips chunks before extraction using a non-LLM score
based on lexical content, entity cues, and embedding novelty. It is used in the
larger-corpus experiments and keeps the same query policy as EffiRAG.

\section{Evaluation}
\label{sec:evaluation}

We first measure the quality--cost trade-off of EffiRAG
and the effect of reducing ingestion work. We then compare EffiRAG with graph
and lower-cost non-graph retrievers. Finally, we examine individual design
choices and test the robustness of the judged-quality conclusion.

\subsection{Evaluation Design and Rationale}

We evaluate on a four-domain sample of UltraDomain~\cite{memorag}: agriculture,
computer science, legal, and mixed. Each domain has five documents and 30
open-ended questions, for 120 queries in total.

The questions require cross-document comparison and corpus-level synthesis.
This makes the benchmark suited to testing whether graph structure helps
connect evidence across documents. We follow LightRAG's public
dataset-construction protocol. Concretely, for each domain we select a fixed
document set and generate open-ended questions from the corpus using an LLM.
The questions target cross-document comparison and synthesis. We use
LightRAG's released prompt and filtering rules without modification.

We organize the comparison by retrieval role. LightRAG-hybrid is the primary
baseline because it combines graph and vector retrieval. PathRAG represents an
alternative graph-retrieval design. NaiveRAG, HyDE, and BM25-vector are
lower-cost non-graph controls. HippoRAG2 is evaluated separately on
gold-answer multi-hop QA, outside the main UltraDomain comparison.

We use DeepSeek-V4-Flash as the answer model
(\texttt{deepseek-v4-flash}, non-thinking mode)~\cite{deepseek-v4-flash}.
It provides a competitive balance of generation quality and inference cost,
and was the model available in our deployment environment.

We use \texttt{all-MiniLM-L6-v2} for embeddings~\cite{sbert,minilm-model-card}.
It runs locally, which keeps embedding cost separate from the provider-LLM
cost accounting. It is also publicly available and stable across runs.
Holding these models fixed isolates the effects of retrieval and graph
construction. System cost includes provider-LLM calls during ingestion and
querying. Judge calls are excluded, and local embeddings are treated as
unbilled.


\begin{table}[t]
\centering
\footnotesize
\caption{Context-budget sensitivity over all 120 queries. Win rate compares
each budgeted setting with full-context generation.}
\label{tab:budget-sensitivity}
\begin{tabular}{rrr}
\toprule
$B$ & Avg.\ tokens & Win rate \\
\midrule
1500 & 1231 & 0.208 \\
2000 & 1694 & 0.317 \\
3000 & 2442 & 0.367 \\
4500 & 3147 & 0.375 \\
\bottomrule
\end{tabular}
\end{table}

We evaluated context budgets of
$B\in\{1500,2000,3000,4500\}$ over all 120 queries
(Table~\ref{tab:budget-sensitivity}).
Increasing $B$ from 3000 to 4500 added 705 selected tokens on average,
while the win rate relative to full-context generation increased by only
0.008. We therefore use $B=3000$ as a practical fixed budget rather than
claiming that it is optimal.

We use pairwise LLM judging. Given the same question and two candidate
answers, GPT-4o-mini~\cite{gpt4o-mini} selects the better answer. It judges
correctness, relevance, and completeness.

Presentation order can affect pairwise judgments~\cite{llmjudge}. Each pair is
therefore judged twice with the answer order reversed. A query is a win for a
system when both orders prefer it. We call a query with no consistent
preference a \emph{split}. A split happens when the two orders disagree, or the
judge returns a tie. We report query-level wins, losses, and splits. When a
single score is needed, splits count as half:
\[
\mathrm{score}
=
\frac{\mathrm{wins}+0.5\,\mathrm{splits}}{N}.
\]

\subsection{Main Quality--Cost Result}

\begin{figure}
\centering
\includegraphics[width=\columnwidth]{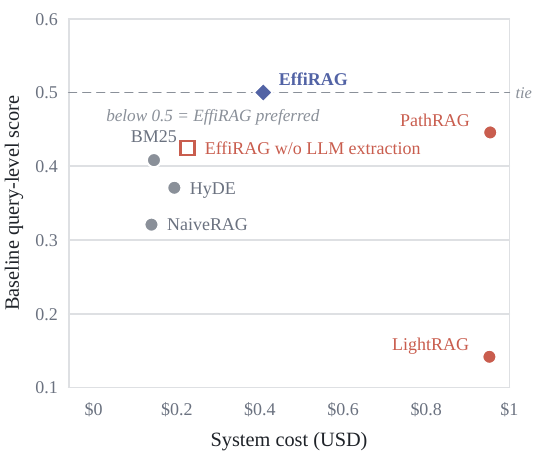}
\Description{Scatter plot of system cost and pairwise quality on UltraDomain.
Each point compares a baseline with EffiRAG. LightRAG-hybrid is both more
expensive and less preferred than EffiRAG.}
\caption{Cost and judged quality on UltraDomain.}
\label{fig:pareto}
\end{figure}

No measured baseline is both cheaper than EffiRAG and preferred over it on
UltraDomain (Figure~\ref{fig:pareto} and Table~\ref{tab:main-pareto}).

\begin{table*}[t]
\centering
\caption{UltraDomain quality and system cost (120 queries). Quality reports
query-level EffiRAG wins / baseline wins / splits. Cost includes provider-LLM
ingestion and querying. Cost delta is relative to EffiRAG. Query time is
cumulative wall-clock time for all 120 queries.}
\label{tab:main-pareto}
\begin{tabular}{lrrrr}
\toprule
System & EffiRAG wins / baseline wins / splits & Cost & Cost delta & Query time (s) \\
\midrule
EffiRAG & (baseline) & \$0.408 & +0.0\% & 1162.5 \\
LightRAG-hybrid & 93/7/20 & \$0.952 & +133.1\% & 888.1 \\
PathRAG & 39/26/55 & \$0.954 & +133.5\% & 1565.3 \\
NaiveRAG & 58/15/47 & \$0.139 & -66.0\% & 1011.6 \\
HyDE & 51/20/49 & \$0.194 & -52.4\% & 1419.4 \\
BM25-vector & 43/21/56 & \$0.145 & -64.6\% & 1059.8 \\
EffiRAG w/o LLM extraction & 44/26/50 & \$0.226 & -44.7\% & 1069.1 \\
\bottomrule
\end{tabular}
\end{table*}

Against LightRAG-hybrid, EffiRAG wins 93 queries, loses 7, and splits 20.
The result is consistent across all four domains: the win/loss margins are
20/2 in agriculture, 27/1 in computer science, 26/1 in legal, and 20/3 in
mixed. Across ingestion and the 120 queries, EffiRAG costs \$0.408, compared
with \$0.952 for LightRAG-hybrid, a reduction of 57.1\%. In this experiment, EffiRAG has
higher total query wall-clock time (1163 versus 888 seconds).

\subsection{Ingestion Cost Control}

EffiRAG extracts every source chunk in the main five-document setting. We ask
whether extracting fewer chunks can reduce cost while keeping the same quality
conclusion.

Random removal could confound extraction volume with content coverage. We
therefore construct a series of extraction settings with progressively smaller
extracted corpora while preserving broad corpus coverage. To do so, we use
a fixed non-LLM selector that estimates the coverage contributed by each chunk.

The selector combines four complementary signals: noun-phrase coverage,
cross-document coverage, corpus representativeness, and expected query-workload
coverage. It uses these signals to estimate the additional coverage contributed
by each chunk per token. The selector is fixed across all settings and is used
only for this sensitivity analysis.

We evaluated $\tau\in\{0,0.03,0.05,0.07,0.10\}$, which retained 68, 45, 24,
4, and 1 of the 68 chunks, respectively. Each setting was evaluated on the same
UltraDomain queries to measure its quality--cost trade-off. As extraction
volume decreased, measured system cost also decreased, but judged answer
quality consistently declined. We therefore retain all-chunk extraction in the
main five-document EffiRAG configuration.

As the boundary of this analysis, we also remove LLM extraction entirely.
\emph{EffiRAG w/o LLM extraction} uses non-LLM noun-phrase extraction while
retaining the same query pipeline.
It costs \$0.226, 44.7\% less than EffiRAG, but EffiRAG wins 44 queries, the
zero-extraction setting wins 26, and 50 are splits.

These results support all-chunk extraction at the five-document scale. As the
corpus grows, however, extracting every chunk becomes increasingly expensive.
The selector above is used only to study the effect of reducing extraction
volume. For larger corpora, we instead evaluate a lightweight online filtering
strategy intended to control ingestion cost. This configuration is
\emph{EffiRAG w/ salience filtering}.


The filter combines lexical cues, entity cues, and embedding novelty to estimate chunk salience.
Chunks with low salience are skipped before graph extraction.

EffiRAG w/ salience filtering remains preferred over LightRAG-hybrid. Its
query-level scores are 0.800 and 0.771 at 10 and 20 documents per domain,
respectively, while costing 4.2$\times$ and 4.5$\times$ less (\$2.316 versus
\$9.799 and \$3.238 versus \$14.608).
EffiRAG w/ salience filtering remains effective at both measured corpus sizes.

\begin{table}[t]
\centering
\footnotesize
\caption{EffiRAG w/ salience filtering versus LightRAG-hybrid at 10 and 20 documents per
domain (120 queries per scale): query-level wins, losses, splits, and score.
$^\dagger$Agriculture has 12 unique documents at the 20-document point.}
\label{tab:scale-lightrag}
\begin{tabular}{lrrrr}
\toprule
Domain & EffiRAG & LightRAG & Split & Score \\
\midrule
\multicolumn{5}{l}{\textit{10 docs per domain}} \\
Agriculture & 21 & 1 & 8 & 0.833 \\
CS & 16 & 3 & 11 & 0.717 \\
Legal & 26 & 1 & 3 & 0.917 \\
Mixed & 19 & 5 & 6 & 0.733 \\
All & 82 & 10 & 28 & 0.800 \\
\midrule
\multicolumn{5}{l}{\textit{20 docs per domain}} \\
Agriculture$^\dagger$ & 18 & 2 & 10 & 0.767 \\
CS & 16 & 4 & 10 & 0.700 \\
Legal & 23 & 0 & 7 & 0.883 \\
Mixed & 19 & 5 & 6 & 0.733 \\
All & 76 & 11 & 33 & 0.771 \\
\bottomrule
\end{tabular}

\vspace{0.5em}
\footnotesize Pooled system cost: EffiRAG w/ salience filtering \$2.316 vs.\ LightRAG-hybrid
\$9.799 at 10 docs (4.2$\times$ lower), and \$3.238 vs.\ \$14.608 at 20 docs
(4.5$\times$ lower).
\end{table}

\subsection{Comparison Scope}

Against PathRAG, EffiRAG reduces cost from \$0.954 to \$0.408. Judged quality
stays close: EffiRAG wins 39 queries, PathRAG wins 26, and 55 are splits.

NaiveRAG, HyDE, and BM25-vector cost less than EffiRAG. Their query-level
pairwise scores against EffiRAG are 0.321, 0.371, and 0.408, respectively
(score = (baseline wins $+\,0.5\times$ splits)$/120$; below 0.5 means EffiRAG
is preferred).

UltraDomain emphasizes retrieving and synthesizing evidence distributed across
documents. Gold-answer multi-hop QA instead emphasizes reasoning over linked
facts to produce an exact answer. We therefore include a complementary multi-hop evaluation using exact-match (EM) and token-level F1.

We use fixed 100-question development slices of HotpotQA~\cite{hotpotqa},
2WikiMultihopQA~\cite{twowiki}, and MuSiQue~\cite{musique}. The pairwise judge
prefers HippoRAG2 in 312 of 600 order-level decisions, and thus EffiRAG's
multi-hop configuration in 288. Their aggregate EM/F1 scores are 0.460/0.583
for EffiRAG and 0.447/0.570 for HippoRAG2. 
The two systems achieve similar aggregate quality under different metrics,
while EffiRAG has substantially lower measured answer-generation cost.

\subsection{Design Evidence}
\label{sec:eval-drivers}

The testbed includes several mechanisms (salience skipping, extractor-tier
routing, entity-repeat hints, query expansion, query-class-dependent retrieval,
and context compression).

Source-chunk grounding has the clearest positive effect. Removing linked
source chunks reduces cost by 49.9\%. The testbed is preferred in 195 of 240
order-level decisions, compared with 44 for the ablated system and one tie
(Table~\ref{tab:ablation}).

Context compression also reduces cost but lowers quality. The testbed is
preferred over the compressed variant by 170/70/0. The remaining rows show
small or inconsistent quality differences within the testbed.

\begin{table*}[t]
\centering
\caption{Mechanism analysis using the separate ablation testbed on
UltraDomain (120 queries). The testbed costs \$0.469. Quality is testbed wins /
variant wins / ties across 240 order-level decisions. Each row changes one
testbed component and is not a direct variant of the main EffiRAG setting.}
\label{tab:ablation}
\begin{tabular}{lrrrr}
\toprule
Ablation & Quality & Cost & Cost delta (vs. testbed) & Avg. context chars \\
\midrule
w/o source chunks & 195/44/1 & \$0.235 & -49.9\% & 3846 \\
w/ compression & 170/70/0 & \$0.259 & -44.8\% & 31209 \\
w/o salience filtering & 116/124/0 & \$0.451 & -3.9\% & 31756 \\
lightweight extraction only & 115/125/0 & \$0.453 & -3.4\% & 32106 \\
w/o entity-repeat hints & 105/135/0 & \$0.468 & -0.4\% & 31622 \\
w/ fixed retrieval counts & 115/125/0 & \$0.430 & -8.5\% & 26411 \\
w/o query expansion & 119/121/0 & \$0.471 & +0.3\% & 31594 \\
\bottomrule
\end{tabular}
\end{table*}

\subsection{Reliability}

Finally, we test whether the main EffiRAG--LightRAG-hybrid conclusion depends
on answer order, ambiguous questions, answer length, or the judge model.

Among the 120 queries, 20 are splits. A two-sided sign test on the remaining
100 queries gives $p<0.001$. The query-level score is 0.858. The bootstrap 95\%
confidence interval is $[0.804,0.908]$.

Two length controls preserve the conclusion. A re-judge explicitly instructed
not to reward answer length gives EffiRAG a query-level score of 0.883. On the
subset whose answer lengths differ by at most 20\%, EffiRAG retains a
query-level score of 0.682.

As an independent-model check, Gemini reviewed 80 answer pairs. We selected
pairs for which GPT-4o-mini's judgments were close or order-sensitive,
preferred the baseline, or favored the shorter answer.

GPT-4o-mini selects the same answer in both orders for 23 of them. The other 57
have no single direction for agreement. Gemini agrees on 17 of the 23
comparable pairs (74\%). This check is consistent with the conclusion that
EffiRAG is preferred over LightRAG-hybrid.

\section{Discussion and Threats to Validity}
\label{sec:threats}

The results suggest that graph structure is most valuable when it helps
retrieve original evidence, with source chunks remaining the answer material.
Removing source chunks
substantially lowers judged quality, whereas additional routing, budgeting,
and compression mechanisms provide no clear improvement. Different workloads
nevertheless favor different quality--cost trade-offs: lightweight retrievers minimize
cost, while EffiRAG and HippoRAG2 show comparable multi-hop QA quality under
different metrics. The
larger-corpus results provide two measured cost-control points for EffiRAG w/
salience filtering.

The evaluation uses 120 LLM-generated queries from four domains and
LLM-based pairwise judgments. Reversed answer orders, length controls,
statistical tests, and an independent-model check reduce judging effects,
but other datasets, human judgments, and model stacks may produce different
results. Dollar costs also depend on provider prices and exclude local
computation.

\section{Conclusion}

This paper asked when graph structure improves retrieval-augmented generation
enough to justify its cost. EffiRAG addresses this question with a
source-grounded design: extracted entities and relations locate relevant evidence, while the
generator answers from the linked source chunks. On UltraDomain, EffiRAG is
preferred over LightRAG-hybrid on 93 of 120 queries while reducing measured
system cost by 57\%. EffiRAG w/ salience filtering remains preferred at lower
measured cost on the larger corpora studied here.

The broader result is that graph-RAG systems require joint evaluation of
answer quality, graph complexity, ingestion cost, query cost, and the evidence
that reaches the generator. Our results
favor using the minimum graph structure needed to connect relevant source
evidence, then selecting a configuration appropriate to the workload and cost
budget.

\section*{Acknowledgements}
This work has emanated from research jointly funded by Taighde Éireann--Research Ireland under Grant Number 13/RC/2094\_2, and by Genesys Cloud Services, Inc.







\FloatBarrier

\end{document}